\def\year{2022}\relax
\documentclass[letterpaper]{article} 
\usepackage{aaai22}  
\usepackage{times}  
\usepackage{helvet}  
\usepackage{courier}  
\usepackage[hyphens]{url}  
\usepackage{graphicx} 
\def\UrlFont{\rm}  
\usepackage{natbib}  
\usepackage{caption} 
\DeclareCaptionStyle{ruled}{labelfont=normalfont,labelsep=colon,strut=off} 
\usepackage{algorithm}
\usepackage{algorithmic}
\usepackage{amsmath}
\usepackage{mathrsfs}
\usepackage{breqn}
\usepackage{xspace}
\usepackage{amsfonts}

\usepackage{booktabs}

\usepackage{newfloat}
\usepackage{listings}
\floatstyle{ruled}
\newfloat{listing}{tb}{lst}{}
\floatname{listing}{Listing}
\usepackage{multirow}
\title{Self-supervised Spatiotemporal Representation Learning by Exploiting Video Continuity}
\author {
    \normalsize
    Hanwen Liang\textsuperscript{\rm 1}\thanks{Corresponding author},
    Niamul Quader \textsuperscript{\rm 1},
    Zhixiang Chi \textsuperscript{\rm 1},
    Lizhe Chen\textsuperscript{\rm 1},
    Peng Dai\textsuperscript{\rm 1}\footnotemark[1],
    Juwei Lu\textsuperscript{\rm 1},
    Yang Wang\textsuperscript{\rm 1,2},
}
\affiliations {
    \textsuperscript{\rm 1} Huawei Noah’s Ark Lab
    \textsuperscript{\rm 2} University of Manitoba, Canada \\
    \{hanwen.liang, niamul.quader1,
    zhixiang.chi, lizhe.chen, peng.dai, juwei.lu, yang.wang3\}@huawei.com
}

\usepackage{bibentry}

\begin{document}

\maketitle
\begin{abstract}
Recent self-supervised video representation learning methods have found significant success by exploring essential properties of videos, e.g. speed, temporal order, etc.
This work exploits an essential yet under-explored property of videos, the \textit{video continuity}, to obtain supervision signals for self-supervised representation learning.
Specifically, we formulate three novel continuity-related pretext tasks, i.e. continuity justification, discontinuity localization, and missing section approximation, that jointly supervise a shared backbone for video representation learning. 
This self-supervision approach, termed as Continuity Perception Network (CPNet), solves the three tasks altogether and encourages the backbone network to learn local and long-ranged motion and context representations. It outperforms prior arts on multiple downstream tasks, such as action recognition, video retrieval, and action localization.
Additionally, the video continuity can be complementary to other coarse-grained video properties for representation learning, and integrating the proposed pretext task to prior arts can yield much performance gains. 
%
\end{abstract}

\section{Introduction}
\label{sec:intro}
Self-supervised video representation learning has recently received great attention owing to its success in learning informative spatiotemporal features from unlabeled videos.
These methods commonly take inspiration from human's visual understanding system and devise various pretext tasks rooted in certain video attributes, e.g., speed or playback rate~\cite{benaim2020speednet,wang2020self,chen2021rspnet,yao2020video}, arrow of time~\cite{wei2018learning}, motion and appearance statistics~\cite{wang2019self} etc. 
However, these attributes over the input video clips are temporally invariant and coarse-grained. For example, speediness is mostly constant for a given clip instance. 
This limits the methods' potential in extensively exploring the fine-grained features of videos~\cite{wang2021unsupervised}.
To learn both coarse- and fine-grained features within a self-supervision framework, in this work, we exploit an essential yet under-explored property of videos, namely, ``video continuity".


\begin{figure}[t]
\centering
\includegraphics[width=0.48\textwidth]{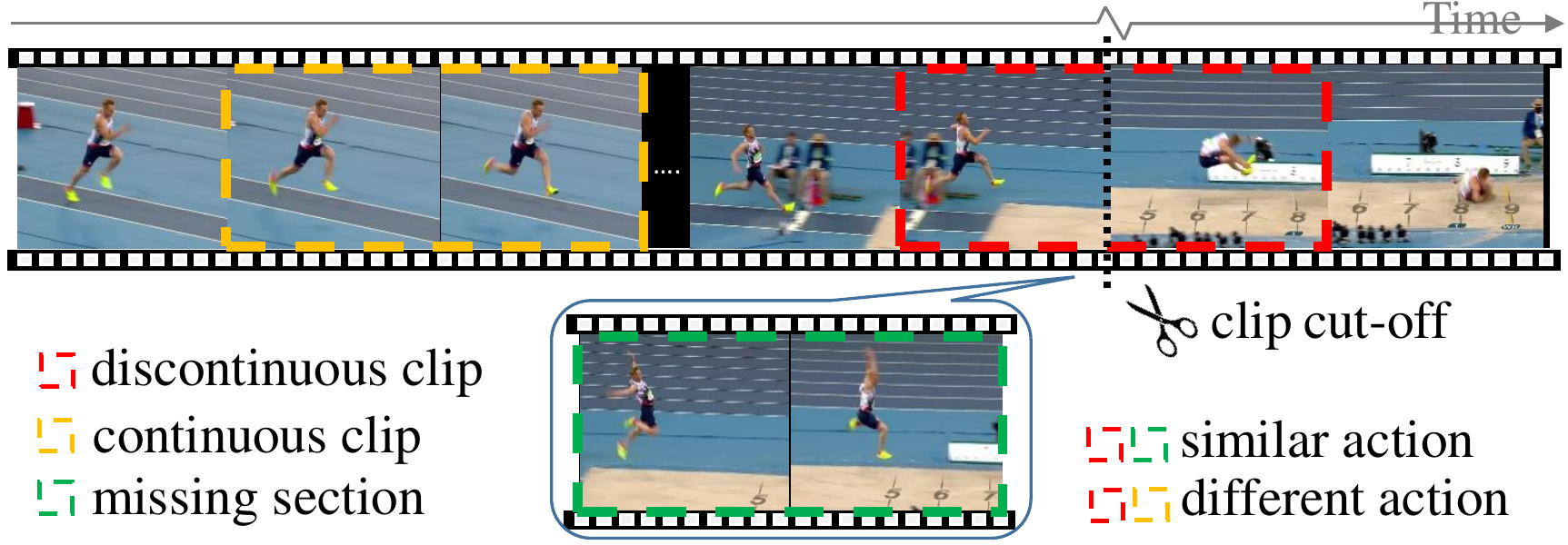}
\caption{Illustration of continuity perception. Observing the long-jump video after manual clip cut-off, a human can easily identify the discontinuity between ``takeoff" and ``landing" and infer the ``action-in-the-air" of the athlete.}
\vspace{-0.5cm}
\label{fig:intro}
\end{figure}

\begin{figure*}[th]
\centering
\includegraphics[width=0.95\textwidth, height=0.41\textwidth]{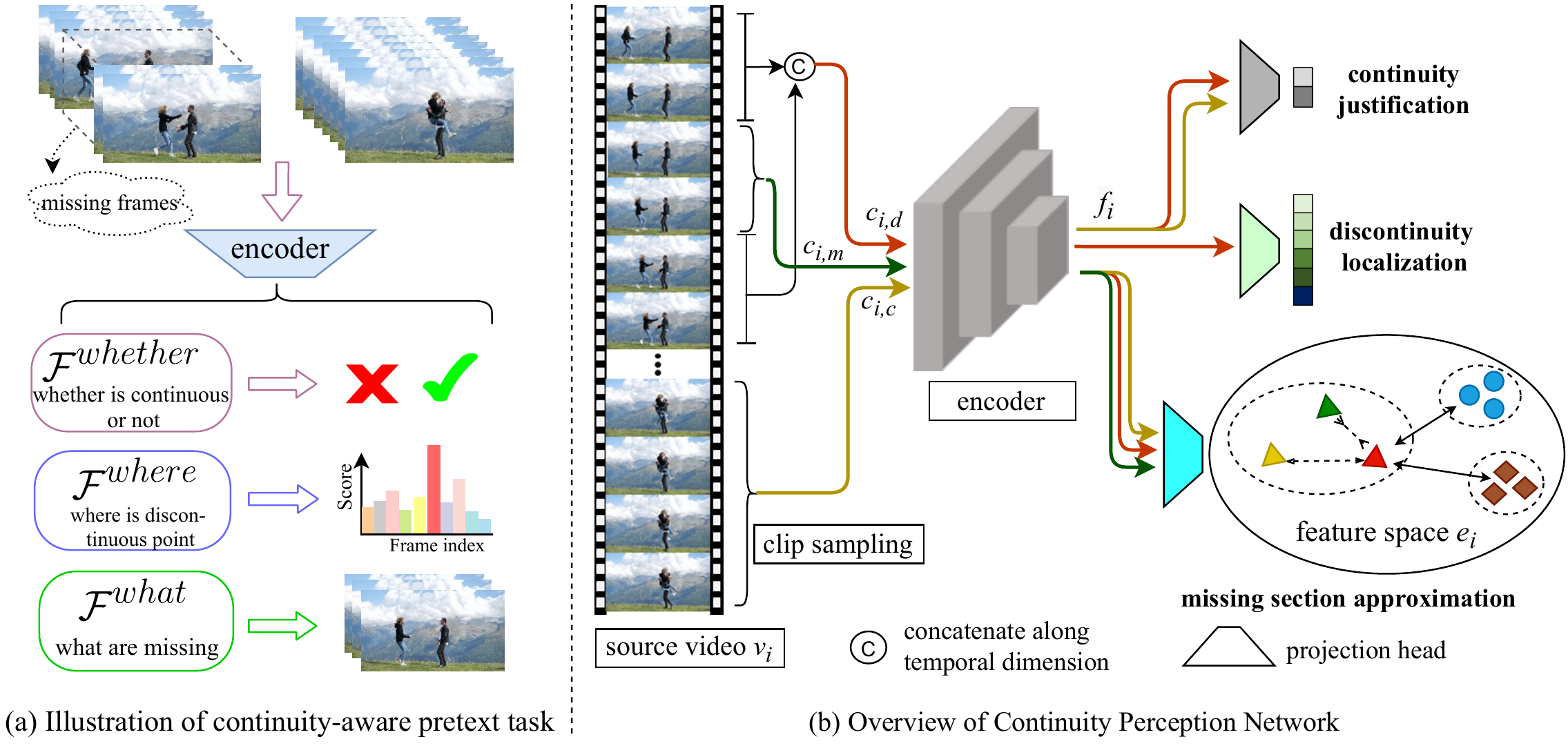} 
\caption{Illustration of the continuity-aware pretext task (a) and the Continuity Perception Network (b). CPNet is composed of a three-branch architecture solving continuity justification, discontinuity localization, and missing section approximation tasks.}
\vspace{-0.2cm}
\label{fig:method}
\end{figure*}

\textit{Video continuity} suggests that objects are represented as the same persisting individuals over time and motion across consecutive frames~\cite{yi2008spatiotemporal}. 
Our choice of using video continuity for designing a self-supervision strategy is motivated by the research findings in cognition sciences~\cite{spelke1995spatiotemporal,yi2008spatiotemporal}. 
They claim that temporal continuity is essential for a correct and persisting understanding of the visual environment. 
In fact, based on years of visual experience, human beings can easily detect discontinuity in videos, if any.
Furthermore, humans are often capable of inferring the high-level semantics associated with the missing section at the discontinuous point. 
For example, in Fig.~\ref{fig:intro}, after we manually cut off some portions from the long-jump video, one can easily notice the discontinuity between ``takeoff" and ``landing" and infer the missing section corresponding to ``action-in-the-air" of the athlete.
We hypothesize that enabling the neural networks to master this exercise of detecting discontinuity and estimating the high-level semantics of the missing sections will empower the model to obtain high-quality spatiotemporal representations of videos. 
This hypothesis is motivated by the following observations.
Effective video embedding requires learning both short- and long-ranged features of videos.
The features could be temporally-rich motion patterns and spatially-rich context information, both of which are complementary to each other~\cite{huang2021self,wang2020self}.
Solving the continuity-aware tasks requires the model to learn those features comprehensively.
Fig.~\ref{fig:method}(a) gives an illustration of the continuity-aware tasks used in this work.
First, identifying \textbf{whether the clip is continuous or not}, i.e. continuity justification, requires a global or long-term view of the motion consistency across the clip.
Inferring a clip to be discontinuous based on a local perception of motion irregularity is insufficient (e.g. a continuous running video could have a local motion irregularity due to the sudden acceleration by the runner).
Second, finding \textbf{where discontinuity occurs}, i.e. discontinuity localization, necessitates a local fine-grained grasp of a dramatic motion change along the video stream.
Third, estimating \textbf{what is missing} semantically, i.e. missing section approximation, requires model to have a high-level understanding of both the motion patterns and the context information in the neighbouring segments.

Following this thread, we propose a Continuity Perception Network (CPNet) solving the novel continuity-aware pretext tasks in Fig.~\ref{fig:method}(a), to learn effective spatiotemporal representations in a self-supervised manner.
We assume there is none or few shot transition in source videos and the discontinuity in the clips refers to the break-point manually created within the same scene (shot). 
Specifically, given the continuous and discontinuous clips, CPNet is trained to finish two discriminative tasks of continuity justification and discontinuity localization, which drive the model to perceive the global and local motion patterns of the video sequence.
For the task of missing content estimation, instead of explicitly reconstruction in RGB space, we formulate it as a contrastive learning task and estimate in the feature space. 
As shown in Fig.~\ref{fig:intro}, since the discontinuous clip encircles its inner missing section in the source video, their motions are more similar to each other than that of two temporally further disjoint clips, even from the same video. 
We first uses a triplet loss~\cite{schroff2015facenet} to pull the features of the discontinuous clip and its inner missing section closer than a disjoint continuous clip from the same video.
Further, based on the observation that clips from the same video have similar appearance compared to those from different videos, we use an additional context-based contrastive loss~\cite{wang2020self,chen2021rspnet} as a regularizer to pull features of clips from the same video together. 
This contrastive learning scheme will promote the features of the discontinuous clip to approximate that of its inner missing section, and encourage the model to learn both fine-grained motion change and context information in the video. 
The CPNet learns video representations by jointly solving these three continuity-aware pretext tasks.

We carry out extensive experiments and demonstrate the superiority of CPNet in learning more effective video representations. CPNet outperforms prior arts on multiple downstream tasks including action recognition, video retrieval and action localization.
Also, the discontinuity localization task is shown to be the most effective pretext task in CPNet, and incorporating it into other typical self-supervised learning methods can bring significant performance gains.

Our major contributions are summarized as follows:
\begin{itemize}
    \item To the best of our knowledge, this is the first work that explicitly exploits video continuity to obtain supervision signals for self-supervised video representation learning.
    \item We propose CPNet to solve the novel continuity-aware pretext tasks and promote the model to learn coarse- and fine-grained motion and context features of the videos.
    \item We conduct comprehensive ablation studies and experiments on multiple downstream tasks to validate the utilities of the proposed CPNet - these include the SOTA or competitive performances on action recognition and video retrieval tasks and evidence of complementary nature to other self-supervised learning methods.
\end{itemize}

\vspace{-0.1cm}
\section{Related Work}

In this section, we discuss two groups of recent advances in self-supervised video representation learning: pretext task-based and contrastive learning-based methods. We also discuss some video analysis techniques that are related to our proposed continuity-aware pretext tasks.

\vspace{-0.1cm}
\subsubsection{Pretext task-based self-supervision.} 
Pretext tasks usually exploit diverse video properties to obtain supervision signals from unlabeled videos.
These tasks operate multiple transformations on source videos for model to recognize and have shown to be effective in self-supervised representation learning~\cite{wang2021removing}.
Examples include identifying temporal order of shuffled clips or frames~\cite{lee2017unsupervised,xu2019self,suzuki2018learning}, predicting video's playback rate~\cite{benaim2020speednet,wang2020self,chen2021rspnet} or motion and appearance statistics~\cite{wang2019self}, identifying the rotation angle of video clips~\cite{jing2018self} or solving spatiotemporal jigsaw puzzles~\cite{ahsan2019video,kim2019self}, etc. 
In this work, we focus on an essential yet less-touched video property, the video continuity.
In some sense, the transformations involved in past works also learn the spatiotemporal continuity implicitly, but they never explicitly use this property for obtaining supervision signals. 
Besides, the transformations in past works are mostly applied at the whole clip or video-level and provide coarse-grained labels for supervision, whereas the proposed method manipulates at a local temporal point (i.e. where the discontinuity occurs). 
This distinction encourages our method to exploit frame-wise temporal coherence and capture more fine-grained motion changes.
Also, our method uses RGB data only and saves the computation cost for the acquisition of cross-modality data or complex hand-crafted visual priors, e.g. dense trajectories~\cite{wang2021unsupervised}.

\vspace{-0.1cm}
\subsubsection{Contrastive learning-based self-supervision.}
Most contrastive learning-based methods build upon the instance discrimination objective and encourage the model to learn some temporal invariance of video instances~\cite{han2019video,han2020memory,wang2020self,han2020self,tao2020selfsupervised,yang2020video,yang2020back}. 
They treat the clips from the same video as positives and those from different videos as negatives.
For example, RSPNet~\cite{chen2021rspnet} designs an A-VID task to learn appearance features of videos with instance discrimination.
Considering that videos have both temporally variant and invariant properties~\cite{dave2021tclr}, CVRL~\cite{qian2021spatiotemporal} relaxes the invariant constraint and develops a weighted temporal sampler to avoid excessive sampling of distant clips.
COCLR~\cite{han2020self} uses cross-modal mining to obtain positive samples across video instances.
Our work integrates both the distinctiveness across different videos and the temporal variance within the same video.
The contrastive learning strategy that takes the discontinuous outer samples and the inner samples as positive pairs, saves the efforts to carefully design complex clip sampler and reduces the computation cost from multiple modalities.

\vspace{-0.1cm}
\subsubsection{Related video analysis techniques.}
Shot boundary detection (SBD) is a task crucial for many video understanding problems. It aims to detect transition and boundaries between consecutive shots~\cite{abdulhussain2018methods}. SBD is similar to our discontinuity localization task in detecting temporal discontinuity. 
The difference is that SBD defines discontinuity at shot-level and tries to grasp the semantic transition between scenes. 
In our proposal, we define discontinuity at frame-level within the same scene and allure the model to obtain more sensitive motion dynamics. 
If needed, SBD techniques, e.g. PySceneDetect tool, can be easily used for pre-processing in our method to obtain source videos with none or few shot transitions.
Video inpainting is another technique related to our pretext tasks, which aims to remove objects or restore missing or tainted regions present in a video sequence~\cite{moran2009video}. 
The similarity between our method and video inpainting lies in explicitly removing and restoring video content based on visual coherence and consistency. 
Differently, video inpainting creates deficiency at spatial dimension and reconstructs the missing areas in RGB space, while our task creates at temporal dimension and implicitly estimate the missing content in feature space.

\vspace{-0.1cm}
\section{Method}

\vspace{-0.3cm}
\subsection{Overview}
\label{method:overview}

Our continuity-perception self-supervision strategy tries to solve multiple continuity-aware pretext tasks altogether to learn effective spatiotemporal video representations.
Let $V$= $\{v_i\}_{i=1}^N$ be an unannotated video set containing N videos. For a clip $c_i$ sampled from video $v_i$, our method aims to learn an encoder $F_{\theta_f}$ parameterized by  $\theta_f$ that maps $c_i$ to continuity-aware rich feature $f_{i}$.
We define feature $f_{i}$ to be rich in continuity-related information if it can be easily used for answering the following – (1) is $c_i$ continuous or not? (2) if $c_i$ is not continuous, where is the discontinuous point? 
(3) if $c_i$ is not continuous, can $f_i$ serve as a good estimation for the feature representation of the missing section in $c_i$?

Fig.~\ref{fig:method}(b) gives an overview of the proposed CPNet, which is a three-branch architecture with all branches sharing the same backbone encoder $F_{\theta_f}$.
Given three non-overlapping clips from video $v_{i}$ – a continuous clip $c_{i,c}$, a discontinuous clip $c_{i,d}$ and its inner missing section $c_{i,m}$, the three branches respectively solve one pretext task and operate jointly to optimize the backbone.
The first branch is used to classify continuous or discontinuous. 
The second branch is used to localize the break-point in $c_{i,d}$. 
The final branch tries to learn appropriate feature representation of the discontinuous clip $c_{i,d}$ that can be a good approximation to the feature representation of $c_{i,m}$ with a contrastive learning scheme. 

\vspace{-0.15cm}
\subsection{Continuity Perception Task Preparation}
\vspace{-0.05cm}
\subsubsection{Data and supervision acquisition.} 
To perform the proposed continuity-aware pretext task, we assume there is only one break-point in timeline to form the discontinuous clip. We use the $l_n$ to denote the length of the discontinuous clip and $l_m$ to denote that of its missing section.
Given a video $v_i$, we first sample an initial clip with length $l_n+l_m$, and then uniformly sample a start-breaking index $j_i \in [1, l_n-1]$ (index starting from 0). 
From the $j^{th}_i$ frame, we extract a clip with length $l_m$ from the initial clip to form the missing section $c_{i,m}$. 
The remaining front and back parts are concatenated to form a discontinuous clip $c_{i,d}$ of length $l_n$. 
We constrain $j_i$ to be within the range $[1, l_n-1]$, so that the missing section generation coincides with the creation of a discontinuous clip. 
We formulate the discontinuity localization as a \textit{($l_n$-1)}-class classification problem with $j_i$ as the label.
The continuous clip, $c_{i,c}$, is formed by randomly sampling a non-overlapping clip with length $l_n$ from a same video. 

\vspace{-0.1cm}
\subsubsection{Model structure.}
We use a typical 3D-ConvNet as the backbone $F_{\theta_f}$ shared by the three pretext tasks.
Each of the tasks respectively incorporates a small projection network at head, denoted as $J_{\theta_j}$, $L_{\theta_l}$  and $E_{\theta_e}$, to process video embeddings $f_i$ from backbone.
All the projection heads have the spatial-temporal average pooling layers, which ensure that the deep feature embeddings of the inputs have the same dimension even when the temporal lengths (i.e. $l_n$ and $l_m$) may be different.
The continuity justification and discontinuity localization tasks use fully-connected layers at the end of project heads ($J_{\theta_j}$ and $L_{\theta_l}$) for classification.
The missing section approximation task uses $E_{\theta_e}$ to embed $f_{{i,c}}$, $f_{{i,d}}$ and $f_{{i,m}}$ to lower-dimensional features $e_{{i,c}}$, $e_{{i,d}}$ and $e_{{i,m}}$.

\vspace{-0.1cm}
\subsection{Continuity Perception Self-supervision}
\subsubsection{Continuity justification.}
For this task, we adopt the same-batch training strategy~\cite{benaim2020speednet} during training, where each batch contains both continuous and discontinuous clips from each video sample. 
The motivation is that, as the positives and negatives are from the same videos, the model will rely on perceiving global motion consistency of the input clips rather than context information or any other artificial cues to solve the task. 
We use the cross-entropy loss ($\mathscr{L}_{CE}$) for the optimization.
Assume that there are $K$ video samples in one batch, the continuity justification loss $\mathscr{L}_{J}$ is,

\vspace{-0.15cm}
\begin{dmath}
\mathscr{L}_{J}=\frac{1}{K}\sum_{i=1}^{K}(\mathscr{L}_{CE}(J_{\theta_{j}}(F_{\theta_{f}}(c_{i,d}))_{1})+\mathscr{L}_{CE}(J_{\theta_{j}}(F_{\theta_{f}}(c_{i,c}))_{0}))
\label{eq:lj}
\end{dmath}
\vspace{-0.15cm} 

\vspace{-0.08cm}
\subsubsection{Discontinuity localization.}
In this task, the label for the discontinuity localization in $c_{i,d}$ is more fine-grained compared to the binary label in the continuity justification task.
It drives the network to perceive more sensitive and fine-grained motion changes and complements the representation learning of the shared backbone $F_{\theta_f}$. 
We also use cross-entropy loss for optimization.
The above two pretext tasks encourage the backbone to learn both coarse- and fine-grained motion patterns. 
The discontinuity localization loss $\mathscr{L}_{L}$ is formulated as,

\vspace{-0.05cm}
\begin{dmath}
\mathscr{L}_{L}=\frac{1}{K}\sum_{i=1}^{K}\mathscr{L}_{CE}(L_{\theta_{l}}(F_{\theta_{f}}(c_{i,d}))_{j_i})
\label{eq:ll}
\end{dmath}
\vspace{-0.15cm}

\vspace{-0.08cm}
\subsubsection{Missing section approximation.}
To further drive $F_{\theta_f}$ to estimate the content in the missing section, we design the following contrastive learning mechanism to obtain an appropriate feature of a discontinuous clip, which can be a good approximation to the feature of its missing section. 
Since the missing section $c_{i,m}$ is surrounded by discontinuous clip $c_{i,d}$, they are temporally connected and supposed to contain more similar motions compared to a further clip even from the same video.
We first takes $c_{i,d}$ as the anchor, $c_{i,m}$ as the positive and $c_{i,c}$ as the negative in a triplet loss (first term in Eq.3)~\cite{schroff2015facenet}, to learn the motion variance within the same video. 
Further, considering that $c_{i,d}$ and $c_{i,c}$ are from the same video and they have more similar context compared to those from different videos, we propose an additional constraint to regularize the feature representations of $c_{i,d}$ and $c_{i,c}$ not to be too far away. 
We adopt the vanilla context-based contrastive learning~\cite{wang2020self} that takes $c_{i,d}$ as the anchor, $c_{i,c}$ as the positive and clips from different videos $\{v_{j}\}_{i\neq j}$ as the negatives in a contrastive loss (second term in Eq.3)~\cite{chen2020simple}. 
Overall, our contrastive continuity learning loss $\mathscr{L}_{E}$ is formulated as:

\vspace{-0.2cm}
\begin{dmath}
\mathscr{L}_{E}=\frac{1}{K}\sum_{i=1}^{K}(\omega \times max(0, \gamma - ({p_i}^{+}-{p_i}^{-}))-(1-\omega)log \frac{q^{+}_i}{{q_i}^{+}+\sum_{j=1,j\neq i}^{K}{q_{i,j}}^{-}})
\label{eq:ccl}
\end{dmath}
\vspace{-0.1cm}

\noindent where $p^{+}_i$= $sim(e_{i,d},e_{i,m})$ and $p^{-}_i$= $sim(e_{i,d},e_{i,c})$ are similarities between positive and negative pairs of the triplet loss, ${q_i}^{+}$= $exp(sim(e_{i,d},e_{i,c})/{\tau})$ is the single positive pair and ${q_{i,j}}^{-}$= $exp(sim(e_{i,d},e_{j,d})/{\tau})+exp(sim(e_{i,d},e_{j,c})/{\tau})$ are similarities between negative pairs in the contrastive loss, $\tau$ is a temperature value which affects the concentration level of feature distributions.
We use cosine similarity for $sim(\cdot,\cdot)$.
$\omega \in [0,1]$ is a hyperparameter to balance the relative contribution of the triplet loss and the contrastive loss. Increasing $\omega$ will promote the model to focus more on within-video motion variance to separate the feature representations of $c_{i,d}$ from $c_{i,c}$ (compared to that of $c_{i,m}$), while decreasing $\omega$ will encourage the model to focus more on cross-video context difference to discriminate video instances.
The combination of the two objectives further pushes the features of $c_{i,d}$ and $c_{i,m}$ to be close to each other, i.e. the feature of the discontinuous clip becomes a good approximation to the feature of its missing pair.
Overall, the model is motivated to learn the context features, including background and object appearance, and fine-grained motion changes in the process.


\vspace{-0.15cm}
\subsubsection{Multi-task joint optimization.} 
Our continuity perception self-supervision strategy trains the shared backbone $F_{\theta_f}$ to generate feature representations that are reliable for each of the above pretext tasks. 
We achieve this by jointly optimizing our network with the multi-loss function,
\begin{equation}
    \mathscr{L} = w_1 \times \mathscr{L}_{J}+w_2 \times \mathscr{L}_{L} + w_3\times \mathscr{L}_{E}
    \label{eq:joint-optimization}
\end{equation}

\noindent where $w_1, w_2, w_3 \in [0,1]$ are the individual weights on the losses. 
The joint optimization promotes the model to learn continuity-aware rich features that embed the global-local motion patterns and the context information of the video.

\vspace{-0.1cm}
\section{Experiments and Results}
\label{sec:exp_set}

\vspace{-0.08cm}
\subsection{Implementation Details}
\vspace{-0.08cm}
\subsubsection{Datasets and architecture.}
We use the following benchmark datasets to evaluate the efficacy of CPNet, i.e. UCF101~\cite{soomro2012ucf101}, HMDB51~\cite{kuehne2011hmdb}, Kinetics400 (abbr. K400)~\cite{kay2017kinetics}, Diving48~\cite{li2018resound} and ActivityNet-v1.3~\cite{caba2015activitynet}.
For UCF101 and HMDB51, We use the training/testing split 1 for fair comparison to prior works.
\textit{ActivityNet-v1.3} is a benchmark dataset for action temporal localization task. 
We use four well-known 3D-ConvNets as the backbone encoder, i.e. C3D~\cite{tran2015learning}, R3D-18(R3D)~\cite{hara2018can}, R(2+1)D-18(R(2+1)D)~\cite{tran2018closer} and I3D~\cite{carreira2017quo}. We add three light-weight projection networks (details in Sec.B in Supp. file) in parallel on top of the video encoder to perform the proposed pretext tasks. 

\vspace{-0.15cm}

\subsubsection{Self-supervised pretraining.} 
In model pretraining stage, we use the training set of UCF101 or K400 without any annotations.
Stochastic gradient descent(SGD) is used for optimization with an initial learning rate of 0.01.
For UCF101 (K400), the model is pretrained with a batch size of 32 (64) for 200 (40) epochs, and the learning rate is decayed by 0.1 at the 100th and 150th (20th and 30th) epoch when the loss plateaus.
We let $\omega$=0.5 in~\eqref{eq:ccl} and $w_1$=$w_2$=1.0, $w_3$=0.1 in~\eqref{eq:joint-optimization}.
For all self-supervised pretraining and downstream tasks, we set the length of input video clip $l_n$ as 16 with a resolution of $112 \times 112$.
The length of the missing section ($l_m$) during pretraining is determined in ablation study.
When pretraining with UCF101(K400), we use 25-fps (15-fps) source frames for both pretrained and evaluated datasets.
More discussion about the choice of fps is in Supp. file.
Common augmentations are applied on input video clips, including color jittering, random scaling, cropping and horizontal flipping.

\vspace{-0.15cm}
\subsubsection{Downstream tasks.}
We use several downstream tasks to evaluate the effectiveness of our method.
\textit{(1) Action recognition:} We append an adaptive spatial-temporal average pooling layer and a fully-connected linear after the pretrained backbone for action recognition.
The model is finetuned end-to-end on the training set of UCF101, HMDB51 or Diving48 for 200 epochs with a batch size of 16. 
We optimize with SGD with an initial learning rate of 0.01, which is decayed by 0.1 at the 80th and 160th epoch when the loss plateaus. 
Following the common evaluation protocol~\cite{wang2020self, wang2021removing}, we perform center cropping and average the scores on 10 uniformly sampled clips on validation splits of these datasets. 
Top-1 accuracy is used as the evaluation metric.
\textit{(2) Video retrieval:} We pretrain the backbone on the training split of UCF101 and add an adaptive spatial-temporal average pooling layer at top to obtain features of video clips.
we perform center cropping and average the features on 10 uniformly sampled clips for each video sample. 
The features of the test set are used to query the k-nearest neighbor videos from the training set. 
Cosine distance is utilized as the similarity metric. 
When the class of a query video appears in the classes of k-nearest training videos, it is considered to be a correct prediction.
We evaluate our method on UCF101 and HMDB51 and use recall at top-k (R@k) as evaluation metric.
\textit{(3) Action temporal localization:} To show the efficacy of our method in learning more fine-grained motion features, we also perform the task of action localization on the ActivityNet-v1.3 dataset.
This task aims to generate action proposals that cover groundtruth instances with high recall and temporal overlap.
We pretrain model on the training set of ActivityNet-v1.3 with the same settings as we pretrain on UCF101, and extract features of both training and validation sets with pretrained backbone.
We adopt the popular BMN~\cite{lin2019bmn} framework for action localization.
BMN is trained on the extracted features of the training set and then evaluated on the validation set.
We adopt Average Recall (AR) and calculate AR under different Average Number of proposals (AN) as AR@AN as evaluation metric.
We also calculate the Area under the AR vs. AN curve (AUC) with AN varying from 0 to 100.

\vspace{-0.2cm}
\begin{table}[!htpb]
\setlength{\tabcolsep}{2.5pt}
\scriptsize
\centering
\caption{Ablation studies of CPNet on action recognition task - its sensitivity to the length of missing section ($l_m$), the individual contribution of each pretext task. ``CJ", ``DL" and ``MSA" respectively denotes continuity justification, discontinuity localization and missing section approximation task. ``CPNet-" denotes not using triplet loss in CPNet.}
\label{tab:tableAblation}
\begin{tabular}{ccccc}
\hline
\multirow{2}{*}{Method} & Length of & \multicolumn{3}{c}{Action recognition} \\
		\cline{3-5} 
		& $l_m$ & UCF101 & HMDB51 & Diving48 \\
\hline \hline
Random Init & -- & 70.3 & 37.5 & 62.7 \\
\hline
& $4$ & {70.9} & 38.1  & 65.4 \\
\multirow{1}{*}{\parbox{1.6cm}{\centering CJ}}  & $8$ & \textbf{74.3} & \textbf{43.6} & \textbf{67.5} \\
 & $16$ & 73.1 & 40.1 & 65.6 \\
  \hline

\multirow{3}{*}{\parbox{1.6cm}{\centering DL}} & $4$ & 78.3 & 47.0 & 70.8  \\
 & $8$ & \textbf{78.4} & \textbf{47.2} & \textbf{71.0} \\
 & $16$ & {76.6} & 45.6 & 70.0 \\
 \hline
 
 \multirow{3}{*}{\parbox{1.6cm}{\centering MSA}} & 4 & \textbf{76.1} & 43.8 & 68.1 \\
 & 8 & 75.5 & \textbf{43.8} & \textbf{68.2} \\
 & 16 & 72.0 & 40.7 & 68.0 \\
 \hline
 & $4$ & 80.1 & 49.9 & 72.1 \\
 \multirow{1}{*}{\parbox{1.6cm}{\centering CPNet}} & $8$ & \textbf{80.7} & \textbf{51.8}  & \textbf{72.5} \\
  & $16$ & 79.8 & 47.3 & 71.6 \\
 \hline
 CJ \& DL & \multirow{4}{*}{\parbox{1.6cm}{\centering $8$}} & 78.6 & 47.3 & 71.9 \\
 CJ \& MSA & & 77.0 & 45.1 & 71.8 \\
 DL \& MSA & & 79.3 & 47.6 & 71.9 \\
 CPNet- & & 80.0 & 49.7 & 72.0 \\
 \hline
\end{tabular}
\vspace{-0.35cm}

\end{table}

\subsection{Ablation Study}
To determine the optimal length of the missing section ($l_m$) and analyze the positive effects brought by each pretext task, we conduct ablation studies with R(2+1)D and evaluate with action recognition task on UCF101, HMDB51 and Diving48. 
In this section, we use 90\% of the training split of UCF101 for pretraining.
During evaluation, for each dataset, 90\% of the training set is used for finetuning (the same 90\% pretrained for UCF101) and the rest 10\% is used for testing.

\vspace{-0.15cm}
\subsubsection{Sensitivity to the length of missing section.}
The length of the missing section ($l_m$) is an important setting in CPNet that affects the model's sensitivity to video continuity. A small value of $l_m$ may make the discontinuity too difficult to detect and the pretext task too ambiguous to solve, whereas a large value may make the task too easy and reduce the necessity for model to learn high-quality discriminative representations.
We test the sensitivity of each pretext task to $l_m$ by pretraining with only one of the tasks activated.
As shown in Table~\ref{tab:tableAblation}, $l_m$= 8 gives better results compared to the lower $l_m$= 4 and the higher $l_m$= 16 for continuity justification and discontinuity localization, though $l_m$= 4 gives comparable results for missing section approximation. 
Similar observations are found in~\cite{wang2020self} that the pretext tasks cannot be too simple or too ambiguous to get good performance.
For CPNet with all three tasks, $l_m$= 8 gives the best result, so we use $l_m$= 8 for experiments in the rest of the paper.
We also show the pretraining performance of the model on the pretext tasks in Sec.C in the Supp. file.

\vspace{-0.1cm}
\subsubsection{Effectiveness of pretext tasks.}
From the $2^{nd}$ to $4^{th}$ sections in Table~\ref{tab:tableAblation}, we can see that each pretext task can bring improvements over the random initialization.
Notably, the discontinuity localization task is the most effective one
among others.
Since the discontinuous location is sampled uniformly along the timeline of the input, this task promotes the model to densely capture the motion changes and learn fine-grained representations corresponding to each timestamp of the input.
Also, Fig.~\ref{fig:method} shows that this task is more efficient as it requires processing only one clip from each video sample.
The light-weight property and effective behavior of this task shows its superiority over many competing pretext task-based and contrastive learning-based approaches.
While other proposed tasks are not as potent as the discontinuity localization, they are complementary to each other and provide the best when all of them (i.e. CPNet) are used together (the $5^{th},6^{th}$ sections in Table~\ref{tab:tableAblation}).
We conjecture that, this joint improvement comes from the complementary local and long-ranged motion and contextual feature learning pathways of the individual tasks. 
We also note that the missing section approximation is similar to a generic context-based contrastive learning without the triplet loss. 
Removing triplet loss from CPNet degrades performances (CPNet- in Table~\ref{tab:tableAblation}), suggesting that our missing section approximation strategy in CPNet works better than a generic context-base contrastive learning.

\subsection{Evaluation of Self-supervised Representation}
In this section, we evaluate the representation capability of our self-supervision strategy in multiple downstream tasks.

\begin{table}[t]
    \setlength{\abovecaptionskip}{0pt}
    \setlength{\belowcaptionskip}{0pt}
    \setlength{\tabcolsep}{1.3pt}
	\scriptsize
	\caption{Comparison with SOTA self-supervised approaches on action recognition task with benchmark datasets.}
 	\resizebox{.48\textwidth}{!}{
	\begin{tabular}{cccccc}
		\hline
		Method & Pretrained & Resolution & Backbone & UCF101 & HMDB51  \\
		\hline
		\hline
		Shuffle\cite{ishan2016unsupervised} & UCF101 & 227$\times$227 & CaffeNet & 50.2 & 18.1 \\
        VCOP~\cite{xu2019self} & UCF101 & 112$\times$112 & R(2+1)D & 72.4 & 30.9 \\
        VCP~\cite{luo2020video} & UCF101 & 112$\times$112 & C3D & 68.5 & 32.5 \\
        PacePred~\cite{wang2020self} & UCF101 & 112$\times$112 & R(2+1)D & 75.9 & 35.9 \\
        PRP~\cite{yao2020video} & UCF101 & 112$\times$112 & R(2+1)D & 72.1 & 35 \\
        TempTrans~\citep{jenni2020video} & UCF101 & 112$\times$112 & R(2+1)D & 81.6 & 46.4 \\
        PSPNet~\cite{cho2021self} & UCF101 & 112$\times$112 & R3D & 70.0 & 33.7 \\
        PSPNet~\cite{cho2021self} & UCF101 & 112$\times$112 & R(2+1)D & 74.8 & 36.8 \\
        STS~\cite{wang2021self} & UCF101 & 112$\times$112 & R(2+1)D & 77.8 & 40.7 \\
        \hline
        \textbf{CPNet} & UCF101 & 112$\times$112 & C3D & \textbf{77.5} & \textbf{45.2} \\
        \textbf{CPNet} & UCF101 & 112$\times$112 & R3D  & \textbf{77.2} & \textbf{46.3} \\
        \textbf{CPNet} & UCF101 & 112$\times$112 & R(2+1)D  & \textbf{81.8} & \textbf{51.2} \\
        \hline
        SpeedNet~\cite{benaim2020speednet} & K400 & 224$\times$224 & S3D-G & 81.8 & 48.8 \\
        TempTrans~\cite{jenni2020video} & K400 & 112$\times$112 & C3D & 69.9 & 39.6 \\
        PacePred~\cite{wang2020self} & K400 & 112$\times$112 & R(2+1)D & 77.1 & 36.6 \\
        RSP~\cite{chen2021rspnet} & K400 & 112$\times$112 & R(2+1)D & 81.1 & 44.6 \\
        VideoMoCo~\cite{pan2021videomoco} & K400 & 112$\times$112 & R3D & 74.1 & 43.6 \\
        VideoMoCo~\cite{pan2021videomoco} & K400 & 112$\times$112 & R(2+1)D & 78.7 & 49.2 \\
        \hline
        \textbf{CPNet} & K400 & 112$\times$112 & C3D & \textbf{80.0} & \textbf{50.7} \\
        \textbf{CPNet} & K400 & 112$\times$112 & R3D  & \textbf{80.8} & \textbf{52.8} \\
        \textbf{CPNet} & K400 & 112$\times$112 & R(2+1)D  & \textbf{83.8} & \textbf{57.1} \\
        \hline
	\end{tabular}}
	\label{tab:actionclassification1}
    \vspace{-0.3cm}
\end{table}

\begin{table}[t]
    \setlength{\abovecaptionskip}{0pt}
    \setlength{\belowcaptionskip}{0pt}
    \setlength{\tabcolsep}{2.2pt}
	\scriptsize
	\caption{Comparison with SOTA self-supervised approaches on action recognition with fine-grained dataset Diving48.}
	\centering
	\begin{tabular}{cccc}
		\hline
		Method & Pretrained & Backbone & Accuracy  \\
		\hline
		\hline
		\textit{Supervised} & Sports1M & C3D & 66.5 \\
        TCLR\cite{dave2021tclr} & Diving48 & R3D & 22.9 \\
        MoCo+BE\cite{wang2021removing} & Diving48 & I3D & 58.3 \\
        MoCo+BE\cite{wang2021removing} & K400 & I3D & 62.4 \\
        \hline
        \textbf{CPNet} & Diving48 & I3D & 67.1 \\
        \textbf{CPNet} & UCF101 & R3D & 70.1 \\
        \textbf{CPNet} & UCF101 & C3D & 72.6 \\
        \textbf{CPNet} & UCF101 & R(2+1)D & 72.8 \\
        \textbf{CPNet} & K400 & R(2+1)D & \textbf{73.6} \\
	\hline
	\end{tabular}
	\label{tab:actionclassification2}
    \vspace{-0.3cm}
\end{table}

\vspace{-0.1cm}
\subsubsection{Action recognition.} 
For action recognition, we pretrain on the training set of UCF101 or K400, and finetune on UCF101, HMDB51 and Diving48 datasets.
Table~\ref{tab:actionclassification1} shows that CPNet achieves consistently superior results over all the previous self-supervised methods on both benchmark datasets.
With the same backbone, our models pretrained on UCF101 even outperform RSP~\cite{chen2021rspnet} and VideoMoCo~\cite{pan2021videomoco} pretrained on a larger K400 dataset. 
CPNet also benefits from pretraining on a larger dataset and gets better results.
It outperforms RSP~\cite{chen2021rspnet} by 2.9\% and 12.5\% on UCF101 and HMDB51 with R(2+1)D.

Also, Table~\ref{tab:actionclassification2} suggests our method is effective for fine-grained action recognition. 
CPNet pretrained on Diving48 with I3D outperforms MoCo+BE~\cite{wang2021removing} by 8.8\% under the same setting, and by 4.7\% even pretrained with K400.
Remarkably, CPNet pretrained on UCF101 with C3D outperforms the supervised model pretrained on a larger dataset Sports1M.
Based on ablation study in Table~\ref{tab:tableAblation}, we give the credit to the discontinuity localization and missing section approximation tasks, which enforce the model to pay more attention to the fine-grained motion features.

\begin{table}[t]
    \setlength{\abovecaptionskip}{0pt}
    \setlength{\belowcaptionskip}{0pt}
    \setlength{\tabcolsep}{2.0pt}
	\centering
	\scriptsize
	\caption{Comparison with SOTA self-supervised approaches on video retrieval task with UCF101 and HMDB51 datasets.}
 	\resizebox{.48\textwidth}{!}{
	\begin{tabular}{ccccccc}
		\hline
		{Method}  & \multicolumn{5}{c}{UCF101/HMDB51} \\
		\cline{2-6} 
		(Pretrained on UCF101) & R@1 & R@5 & R@10 & R@20 & R@50 \\
		\hline
		\hline
	  \multicolumn{7}{c}{Backbone: AlexNet}\\
	  \hline
      Jigsaw~\cite{noroozi2016unsupervised} & 19.7/-- & 28.5/-- & 33.5/-- & 40.0/-- & 49.4/-- \\
      OPN~\cite{lee2017unsupervised} & 19.9/-- & 28.7/-- & 34.0/-- & 40.6/-- & 51.6/-- \\
      \hline
      \multicolumn{7}{c}{Backbone: C3D}\\
	  \hline
      VCOP~\cite{xu2019self} & 12.5/7.4 & 29.0/22.6 & 39.0/34.4 & 50.6/48.5 & 66.9/70.1  \\
      VCP~\cite{luo2020video} & 17.3/7.8 & 31.5/23.8 & 42.0/35.5 & 52.6/49.3 & 67.7/71.6 \\ 
      PacePred~\cite{wang2020self} & 31.9/12.5 & 49.7/32.2 & 59.2/45.4 & 68.9/61.0 & 80.2/80.7 \\
      PRP~\cite{yao2020video} & 23.2/10.5 & 38.1/27.2 & 46.0/40.4 & 55.7/56.2 & 68.4/75.9 \\
      \textbf{CPNet} & \textbf{33.2/13.7} & \textbf{50.0/33.2} & \textbf{55.8/45.5} & \textbf{64.5/59.3} & \textbf{76.5/78.7} \\
      \hline
      \multicolumn{7}{c}{Backbone: R3D}\\
	  \hline
      VCOP~\cite{xu2019self} & 14.1/7.6 & 30.3/22.9 & 40.4/34.4 & 51.1/48.8 & 66.5/68.9 & \\
      VCP~\cite{luo2020video} & 18.6/7.6 & 33.6/24.4 & 42.5/36.6 & 53.5/53.6 & 68.1/76.4 & \\
      PacePred~\cite{wang2020self} & 23.8/9.6 & 38.1/26.9 & 46.4/41.1 & 53.5/56.1 & 69.8/76.5 & \\
      PRP~\cite{yao2020video} & 22.8/8.2 & 38.5/25.8 & 46.7/38.5 & 55.2/53.3 & 69.1/75.9 & \\
      TempTrans~\cite{jenni2020video} & 26.1/-- & 48.5/-- & 59.1/-- & 69.6/-- & 82.8/-- \\
      MemDPC~\cite{han2020memory} & 20.2/7.7 & 40.4/25.7 & 52.4/40.6 & 64.7/57.7 & --/-- & \\
      PSPNet~\cite{cho2021self} & 24.6/10.3 & 41.9/26.6 & 51.3/38.8 & 62.7/54.6 & 67.9/76.8 & \\
      \textbf{CPNet} & \textbf{35.1/16.5} & \textbf{49.0/35.5} & \textbf{57.2/47.5} & \textbf{67.3/60.0} & \textbf{76.6/79.2} & \\
      \hline
      \multicolumn{7}{c}{Backbone: R(2+1)D}\\
	  \hline
      VCOP~\cite{xu2019self} & 10.7/5.7 & 25.9/19.5 & 35.4/30.7 & 47.3/45.8 & 63.9/67.0 & \\
      VCP~\cite{luo2020video} & 19.9/6.7 & 33.7/21.3 & 42.0/32.7 & 50.5/49.2 & 64.4/73.3 & \\
      PacePred~\cite{wang2020self} & 25.6/12.9 & 42.7/31.6 & 51.3/43.2 & 61.3/58.0 & 74.0/77.1 & \\
      PRP~\cite{yao2020video} & 20.3/8.2 & 34.0/25.3 & 41.9/36.2 & 51.7/51.0 & 64.2/73.0 & \\
      
      \textbf{CPNet} & \textbf{35.3/14.0} & \textbf{49.9/32.8} & \textbf{58.6/45.8} & \textbf{67.0/60.5} & \textbf{77.8/77.2} & \\
      \hline
	\end{tabular}}
	\label{tab:videoretrieval}
    \vspace{-0.25cm}
\end{table}

\vspace{-0.1cm}
\subsubsection{Video retrieval.} Table~\ref{tab:videoretrieval} presents the results on video retrieval task with model pretrained on the training split of UCF101. 
CPNet achieves competitive results to prior arts with C3D and R3D backbone.
With R(2+1)D, it achieves the highest recall for all K values in UCF101 and HMDB51.
Noticeably, the proposed method outperforms the second-best PacePred~\cite{wang2020self} by 9.7\% when K=1 on UCF101 using R(2+1)D. 

\begin{table}[t]
    \setlength{\abovecaptionskip}{0pt}
    \setlength{\belowcaptionskip}{0pt}
    \setlength{\tabcolsep}{2.2pt}
	\scriptsize
	\caption{Comparison between features from CPNet and other self-supervised approaches for temporal action localization task. Evaluation is performed on validation set of ActivityNet-v1.3 in terms of AR@AN and AUC.}

 	\resizebox{.48\textwidth}{!}{

	\begin{tabular}{cccccc}
		\hline
		Method & AR@1 & AR@5 & AR@10 & AR@100 & AUC  \\
		\hline
		\hline
		Random Init & 0.2837 & 0.3612 & 0.4205 & 0.6636 & 55.917\\
		\hline
		RSP~\cite{chen2021rspnet} & 0.3144 & 0.4211 & 0.4879 & 0.7104 & 60.989\\
		PRP~\cite{yao2020video} & 0.3077 & 0.4041 & 0.4674 & 0.6997 & 60.248\\
		VCOP~\cite{xu2019self} & 0.3024 & 0.3969 & 0.4586 & 0.6907 & 59.451 \\
		TempTrans~\cite{jenni2020video} & 0.3082 & 0.3998 & 0.4612  & 0.6959 & 59.756\\
		\hline
		\textbf{CPNet} & \textbf{0.3205} & \textbf{0.4273} & \textbf{0.4954} & \textbf{0.7133} & \textbf{62.064} \\	\hline
	\end{tabular}}
	\label{tab:actionlocalization}
    \vspace{-0.2cm}
\end{table}

\vspace{-0.1cm}
\subsubsection{Action temporal localization.} 
Action localization aims to generate temporal boundaries for the action instance in untrimmed videos, which requires a thorough and fine-grained video understanding.
We pretrain with C3D using multiple typical self-supervised video representation learning methods on the training set of ActivityNet-v1.3. 
Then we use pretrained models to extract multiple sets of video features of both training and validation dataset.
For each set of features, BMN~\cite{lin2019bmn} framework is first trained on features of the training dataset and evaluated on features of the validation dataset (details in Sec.E of Supp. file). Results are shown in Table~\ref{tab:actionlocalization}.
CPNet exploits local and long-ranged motion patterns and context information, and generate more powerful features, leading to better action temporal localization performance.

\begin{figure}[t]
\centering
\includegraphics[width=0.95\columnwidth,height=0.6\columnwidth]{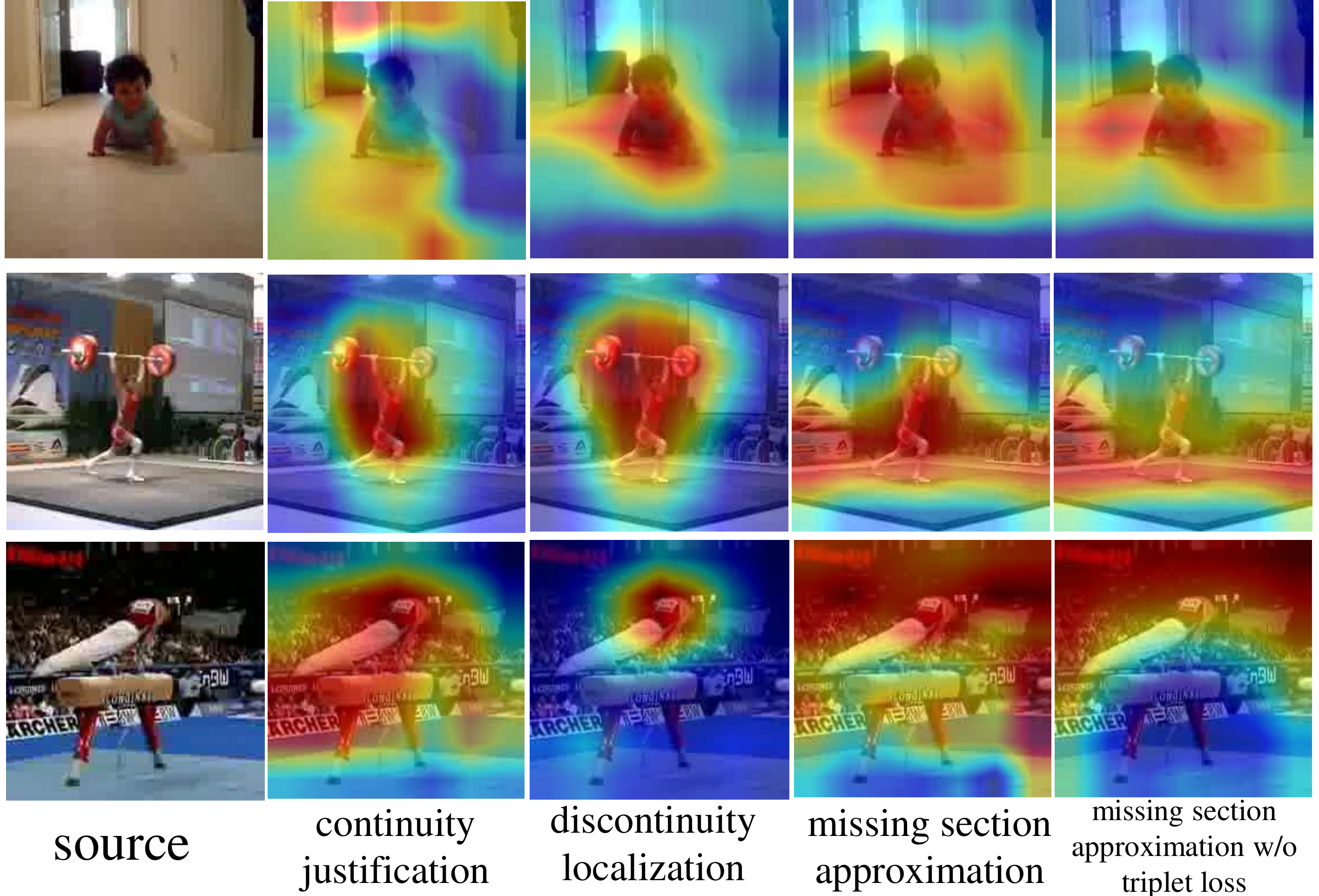}
\caption{Illustration of salient regions for each continuity-aware pretext task.}
\label{fig:cam}
\vspace{-0.4cm}
\end{figure}

\vspace{-0.15cm}
\subsection{Visualization Analysis}
To gain a better understanding of what spatiotemporal clues are learned during pretraining, we visualize the salient regions that contribute the most to the accomplishments of the proposed tasks, with small modifications on CAM~\cite{zhou2016learning,wang2021removing} technique. 
Specifically, we sample some discontinuous clips from the videos and extract three sets of feature maps before average pooling layers in the projection networks of three tasks. 
For the missing section approximation branch, we also extract feature maps when pretraining without triplet loss. 
We average the feature map along the channel dimension and then resize the compressed features along spatial dimensions to the size of original video frames and and overlay on them.
In Fig.~\ref{fig:cam}, the middle frame of each clip is used to visualize heatmaps.
In the first two tasks of continuity justification and discontinuity localization, these examples suggest a strong correlation between highly activated regions and the dominant moving objects in the scene.
The first task tends to focus more on global movements over time, e.g., the crawling path of the baby, the rising curve of the barbell and swinging over the pommel horse.
The second task concentrates more on the dominant moving object and its fine-grained motion changes.
In comparison, the third task of missing section approximation has larger salient regions over important context cues in the video.
We find that removing the triplet loss removes salient regions from the fine-grained motion change areas, suggesting that the triplet loss encourages learning of such features.

\newcommand{\minitab}[2][l]{\begin{tabular}{#1}#2\end{tabular}}

\begin{table}[t]
\setlength{\tabcolsep}{4.1pt}
\scriptsize
\centering
\caption{Action recognition performances by incorporating video continuity into prior methods. ``DL" denotes ``discontinuity localization" task. Models are pretrained on UCF101.}
\label{tab:complementary}
\vspace{-0.3cm}

\begin{tabular}{cccccl}
\hline
\multirow{2}{*}{Backbone} & \multirow{2}{*}{Method} & \multicolumn{2}{c}{Baseline} & \multicolumn{2}{c}{Baseline + DL task} \\ 
\cline{3-6} 
& & \multicolumn{1}{c}{UCF101} & HMDB51 & \multicolumn{1}{c}{UCF101} & HMDB51 \\ 
\hline
\hline

\multirow{3}{*}{\minitab[c]{PacePred\\\tiny{(Wang et al. 2020)}}} & C3D & {76.0} & {42.8} & {76.6(+0.6)} & 44.5(+1.7) \\
 & R3D & {72.6} & {41.3} & {75.4(+2.8)} & 44.6(+3.3) \\
 & R(2+1)D & {75.1} & {47.2} & {79.0(+3.9)} & 49.2(+2.0) \\
  \hline
  
\multirow{3}{*}{\minitab[c]{VCOP\\\tiny{~\cite{xu2019self}}}} & C3D & {76.4} & {45.0} & {77.9(+1.3)} & 48.2(+3.2) \\
 & R3D & 74.8 & 45.8 & {77.0(+2.2)} & 47.9(+2.1) \\
 & R(2+1)D & {78.8} & {48.2} & {80.1(+1.3)} & 48.7(+0.5)  \\
 \hline
 
\multirow{3}{*}{\minitab[c]{MOCO\\\tiny{~\cite{he2020momentum}}}} & C3D & {72.3} & {39.8} & {72.9(+0.6)} & 41.2(+1.4) \\
 & R3D & {71.8} & {41.8} & {74.4(+2.6)} & 46.2(+4.4) \\
 & R(2+1)D & {75.6} & {46.0} & {78.5(+2.9)} & 47.7(+1.7) \\
  \hline
  
\end{tabular}
\vspace{-0.4cm}
\end{table}

\vspace{-0.08cm}
\subsection{Complementary with Existing Methods.}
As we mentioned in Sec.~\ref{sec:intro}, the most distinguishing point between video continuity and the video attributes used in prior works is that video continuity is a more fine-grained and temporally variant attribute.
To show if this attribute can be complementary to other coarse-grained global attributes, i.e. playback rate, temporal order of clips, we further exploit the utility of the discontinuity localization task by integrating it to prior self-supervised learning approaches, i.e. PacePred~\cite{wang2020self}, VCOP~\cite{xu2019self} and MOCO~\cite{he2020momentum} (All self-implemented. Details in Sec.F of Supp. file).
The reason for choosing this pretext task is that it is easy-to-implement and the most effective component of CPNet, and adding it does not bring much computation or memory cost to the base method.
Table~\ref{tab:complementary} shows that the integration consistently brings considerable improvement for action recognition tasks with multiple backbones. 
It suggests that our novel discontinuity localization task can also be a powerful tool used with emerging SOTA self-supervision approaches and that the video continuity attribute can be complementary to other video attributes for spatiotemporal representation learning.


\vspace{-0.1cm}
\section{Conclusion}
This work introduces a novel self-supervised learning framework, CPNet, that explores video continuity property to formulate pretext tasks. 
These continuity-aware pretext tasks, particularly the discontinuity localization task, are easy-to-implement and effective individually in video representation learning. They can also complement prior self-supervision arts for performance gains. 
Using these tasks together within CPNet learns fine- and coarse-grained motion and context features, and leads to competitive or SOTA performances on multiple video understanding tasks.


\bibstyle{aaai}
\bibliography{main}
\clearpage

\appendix
\section{Discussion about Shot Transition}
In our method, we assume that there is none or few shot transition in the source videos, to be precise, hard shot transitions.
Hard shot transition refers to a sudden change in temporal visual information, where the two successive shots are concatenated directly without any smoothing effect~\cite{abdulhussain2018methods}.
The discontinuity from sharp or hard shot transitions in a clip may look similar to that created by manual cut-off in our proposed method, which breaks the motion persistence across frames. This may interfere with solving our proposed continuity-aware pretext tasks. 

In our observation, we find that such shot transitions are rare in the 16-frame clips sampled from any of the datasets we used for pretraining.
The benchmark datasets UCF101, HMDB51, Kinetics400, and Diving48 all contain trimmed short videos without hard shot transition.
Such shot transitions occur slightly more on the ActivityNet-v1.3 dataset. Even on this dataset, our CPNet performs very well, suggesting that CPNet is not sensitive to small occurrences of shot transitions. 
For a video dataset that has large numbers of shot transitions, ensuring a better continuous vs. discontinuous clip sampling can be done with ease by first preprocessing the videos with a shot boundary detection (e.g. PySceneDetect tool~\cite{castellano2020pyscenedetect}) and then sampling within video segments without any shot transition. It is also noteworthy to mention that such shot boundary detection approaches are extremely fast and efficient (e.g. over 1000 fps on CPU with PySceneDetect tool).


\section{Architecture of Projection Network}
In our proposed method, we add three light-weight projection networks in parallel on top of the backbone encoder to perform the continuity-aware pretext tasks. 
Each projection network solve one pretext task, and the architecture details are shown in Table~\ref{tab:pn-architecture}.
Following~\cite{han2020self}, each projection network has a 3D-Conv layer followed by a spatial-temporal adaptive pooling layer.
The ``CJ" and ``DL" tasks use fully-connected layers at the end for classification, and ``MSA" also uses a linear layer to map features into lower  128-dimensions.

\begin{table}[!htbp]
\caption{The architecture and layer specifications for projection network of each pretext task. `CJ', `DL' and `MSA' respectively denotes the task of continuity justification, discontinuity localization and missing section approximation.}
\begin{tabular}{c|c}
\hline
{Task} & {Specifications} \\ 
\hline  
\hline
\multirow{4}{*}{CJ} 
& 3D-Conv, 512 filters, kernel $3$, stride $1$, padding $1$ \\
\cline{2-2}
 & 3D-BatchNorm --- ReLU \\
\cline{2-2}
 & 3D-Adapative-AvgPool 1$\times$1$\times$1 \\
\cline{2-2}
& Linear $512\times 2$ \\
 \hline
\multirow{4}{*}{DL} 
& 3D-Conv, 512 filters, kernel $3$, stride $1$, padding $1$ \\
\cline{2-2}
 & 3D-BatchNorm --- ReLU \\
\cline{2-2}
 & 3D-Adapative-AvgPool 1$\times$1$\times$1 \\
\cline{2-2}
& Linear $512\times (l_n-1)$ \\
\hline
\multirow{4}{*}{MSA} 
& 3D-Conv, 512 filters, kernel $3$, stride $1$, padding $1$ \\
\cline{2-2}
 & 3D-BatchNorm --- ReLU \\
\cline{2-2}
 & 3D-Adapative-AvgPool 1$\times$1$\times$1 \\
\cline{2-2}
& Linear $512\times 128$ \\
\hline
\end{tabular}
\label{tab:pn-architecture}
\end{table} 

\section{Performance on Pretext tasks}
In the main paper, we mostly discuss about how CPNet learns effective video representations and performs on downstream tasks. The performance on the pretext tasks themselves can provide valuable insight into CPNet's optimization process and its generalizability across different datasets.
Following the experimental settings in ablation study, we randomly extract 90\% of training split of UCF101 for pretraining with all the pretext tasks activated, and the rest 10\% is used for validation.

\begin{table}[t]
\setlength{\tabcolsep}{2.2pt}
\scriptsize
\centering
\caption{The performance of CPNet on pretext tasks, i.e. continuity justification and discontinuity localization.
The CPNet is pretrained on UCF101 with all the three pretext tasks activated. 
Here we show the classification results on validation sets from UCF101 and HMDB51 datasets.}

\begin{tabular}{c|c|cc}
\hline
\multirow{2}{*}{Method} & \multirow{2}{*}{Settings} & \multicolumn{2}{c}{Pretext task} \\ 
\cline{3-4} 
& & \multicolumn{1}{c}{UCF101} & HMDB51 \\ 
\hline \hline
\multirow{3}{*}{Continuity justification} & $l_m=4$ & {92.1} & {86.1} \\
 & $l_m=8$ & {95.1} & {90.1} \\
 & $l_m=16$ & {97.5} & {92.2} \\
\hline
\multirow{3}{*}{Discontinuity localization} & $l_m=4$ & {94.8} & {86.3} \\
 & $l_m=8$ & {97.4} & {91.0} \\
 & $l_m=16$ & {99.3} & {93.5} \\
\hline
\hline
\end{tabular}
\label{tab:cj-dl}
\end{table}

From the evaluation results in Table~\ref{tab:cj-dl}, we can see the CPNet performs very well on these two pretext tasks.
The validation accuracy on UCF101 dataset is above 90\% for $l_m$= 4,8 or 16.
To verify the generalizability of model on these tasks, the pretrained model is further evaluated on HMDB51 dataset. 
The results suggest good generalizability of model on the pretext tasks ($>90\%$ $accuracy$ for $l_m=8$). 
We also notice that the pretext task performance increases as the length of missing section $l_m$ increases. This is expected since the discontinuity becomes more apparent when the length of missing section is larger.

\section{Discussion about time-span of video clip}
For a fixed temporal dimension, the practically-used fps of source videos affects the time-span of the video clip.
The smaller the fps is, the longer time-span the video clip covers.
In our experiments, we use K400 and UCF101 for pretraining and UCF101, HMDB51 and Diving48 in action classification task.
When evaluating with Diving48, we always use 25-fps regardless of the pretraining dataset.
As the source videos of UCF101 and HMDB51 are of 25-fps, when pretraining on UCF101, all the pretraining and evaluation datasets are prepared with 25-fps.
As the source videos of K400 are of 10-25 fps, when pretraining with K400, all the pretraining and evaluation datasets are prepared with 15 fps.
The results are shown in Table 2 and Table 3 in the main paper.
For K400, we also evaluate the pretrained model on 25-fps datasets, i.e. pretraining with 15-fps and evaluating with 25-fps, and show the results in Tab.3.
The results suggest that CPNet can benefit from pretraining on a larger dataset and get better results.
Increasing the time-span(lower fps) can further improve the model performance, and the similar conclusion is also shown in ~\cite{feichtenhofer2021large}. 
Actually, the fps used during experiments is rarely reported in prior self-supervised video representation learning works, and we hope our work can encourage other researchers to provide more details about this point.

\begin{table}[t]
    \setlength{\abovecaptionskip}{0pt}
    \setlength{\belowcaptionskip}{0pt}
    \setlength{\tabcolsep}{1.3pt}
	\scriptsize
	\caption{Comparison with different time-span on action recognition task with benchmark datasets. All the experiments use the same clip dimension of 16$\times$112$\times$112.}
 	\resizebox{.48\textwidth}{!}{
	\begin{tabular}{ccccc}
		\hline
		Pretrained & fps(pretrain/eval) & Backbone & UCF101 & HMDB51  \\
		\hline
		\hline
        UCF101 & 25/25 & C3D & 77.5 & 45.2 \\
        UCF101 & 25/25 & R3D  & 77.2 & 46.3 \\
        UCF101 & 25/25 & R(2+1)D  & 81.8 & 51.2 \\
        \hline
        \hline
        K400 & 15/25 & C3D & 78.4 & 47.1 \\
        K400 & 15/25 & R3D  & 78.2 & 47.3 \\
        K400 & 15/25 & R(2+1)D  & 82.1 & 53.2 \\
        \hline
        \hline
        K400 & 15/15 & C3D & \textbf{80.0} & \textbf{50.7} \\
        K400 & 15/15 & R3D  & \textbf{80.8} & \textbf{52.8} \\
        K400 & 15/15 & R(2+1)D  & \textbf{83.8} & \textbf{57.1} \\
        \hline
	\end{tabular}}
	\label{tab:actionclassification-supp}
\end{table}

\section{Training Details with BMN framework}
For action temporal localization task, we pretrain on the training set of AcitivityNet-v1.3 using different self-supervised learning methods with C3D as the backbone encoder.
We add an adaptive spatial-temporal average pooling layer after the pretrained backbones and extract multiple sets of video features of both training and validation set of ActivityNet-v1.3.
For each video, we uniformly set 100 starting points along the temporal dimension, and extract 100 clip-level features for each video sample.

For each set of the features, we train a popular BMN~\cite{lin2019bmn} framework on the features of training dataset to perform action localization.
Following~\cite{lin2019bmn}, the BMN model is optimized with Adam~\cite{kingma2014adam} for 9 epochs with a batch size of 16. The initial learning rate is set to 0.001 and decayed by 0.1 at epoch 7.

\section{Implementation of Method Integration}
To show if the video continuity attribute can be complementary to other coarse-grained global video attributes, we integrate the task of discontinuity localization to some prior self-supervised learning approaches.
We give the details of the integration in this section.

\subsubsection{Integration to PacePred.} 
PacePred~\cite{wang2020self} is one of the first works that exploit speed or playback rate of video to obtain supervision signal for self-supervised video representation learning.
It uses multiple sampling rate to sample frames from source videos and obtains video clips with different paces.
The model is trained to recognize the paces of clips and it also involves context-based contrastive objective to learn appearance features of videos.
Please refer to~\cite{wang2020self} for details.
We try to integrate the discontinuity localization and encourage the model to finish discontinuity localization and the original tasks at the same time.
Specifically, during data preparation, we first sample an initial clip with a certain sampling rate and temporal length $l_n+l_m$.
Then we create a discontinuous clip with length $l_n$ from the initial clip in the same way as we create in CPNet.
We add an auxiliary projection network after the backbone of PacePred and network details are the same as ``DL" shown in Table.~\ref{tab:pn-architecture}.
The modified PacePred is required to solve an additional discontinuity localization task with a cross-entropy loss during pretraining.
Following~\cite{wang2020self}, we use SGD optimizer with an initial learning rate 0.001 and set the weight on the discontinuity localization loss as 1.0.

\subsubsection{Integration to VCOP.} 
VCOP~\cite{xu2019self} leverages the chronological order of videos and learn the spatiotemporal representations by predicting the order of shuffled clips from the video.
Given a video, VCOP samples three non-overlapping clips and shuffles the temporal order of these clips. 
The model first projects the three clips to the feature embeddings and predicts the correct order based on the concatenation of the three feature embeddings.
We also integrate the discontinuity localization task to VCOP.
Specifically, during data preparation, we first sample three non-overlapping initial clips with temporal length $l_n+l_m$ from a source video.
Then we create three discontinuous clips with length $l_n$ from the initial clips in the same way as we create in CPNet.
We add an auxiliary projection network after the backbone of VCOP to solve the discontinuity localization task based on the feature embeddings of the inputs.
The modified VCOP is required to predict the location of discontinuity as well as the order of shuffled clips during pretraining.
We follow experimental settings in ~\cite{xu2019self} and use SGD to optimize the model. 
The weight on the discontinuity localization loss is also set as 1.0.

\subsubsection{Integration to MOCO.}
MOCO~\cite{he2020momentum} proposes a framework of unsupervised image representation learning as a dynamic dictionary look-up.
It follows the spirit of instance discrimination, and uses memory bank and momentum-based update mechanism to take advantage of a large-scale negative set in contrastive learning.
Due to its remarkable performance, this mechanism has been adopted to video domain~\cite{chen2021rspnet,wang2020self,han2020self}. 
We use the \textit{InfoNCE} model proposed in~\cite{han2020self} as baseline and integrate the proposed discontinuity localization task.
Given a clip from a source video as anchor, \textit{InfoNCE} trains the model to discriminate its transformed version against other samples from different source videos, where the transformed versions could be clips from the same source video or other augmented versions of the anchor clip.
To incorporate the discontinuity localization task, we sample initial clips with temporal length $l_n+l_m$ and create discontinuous clips with length $l_n$ from the initial clips as anchors.
Similarly, we add an auxiliary projection network after the feature extractor of \textit{InfoNCE} to solve the discontinuity localization task.
The modified \textit{InfoNCE} is required to predict the location of discontinuity as well as do instance discrimination during pretraining. 
We follow experimental settings in~\cite{han2020self} and use Adam to optimize the model for 200 epochs, with an initial learning rate of 0.001. Weight of the discontinuity localization loss is 1.0.

\begin{figure}[t]
\centering
\includegraphics[width=0.48\textwidth]{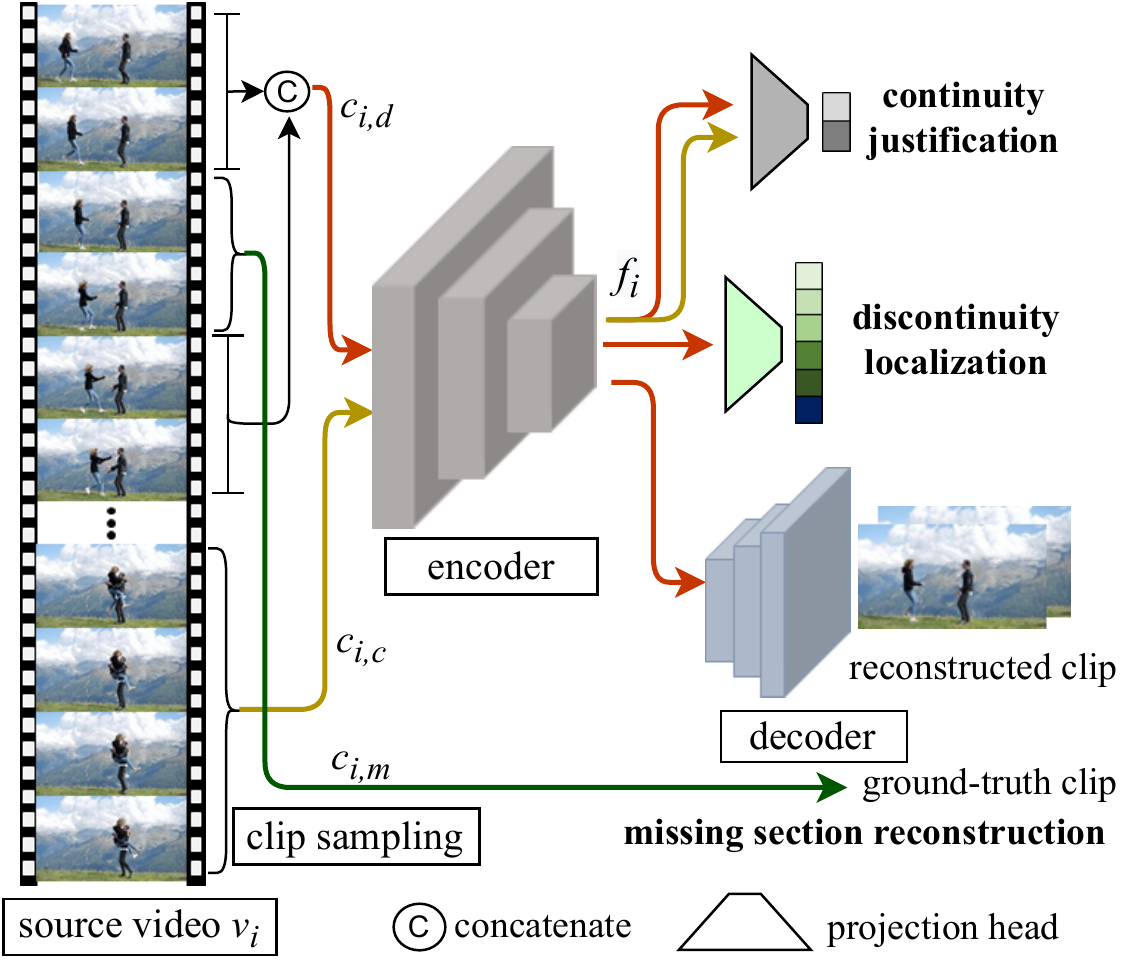}
\caption{Illustration of Continuity Reconstruction Network(CRNet).}
\label{fig:method_supp}
\end{figure}

\begin{figure}[h]
    \centering
    \includegraphics[width=0.98\columnwidth,height=0.2\textwidth]{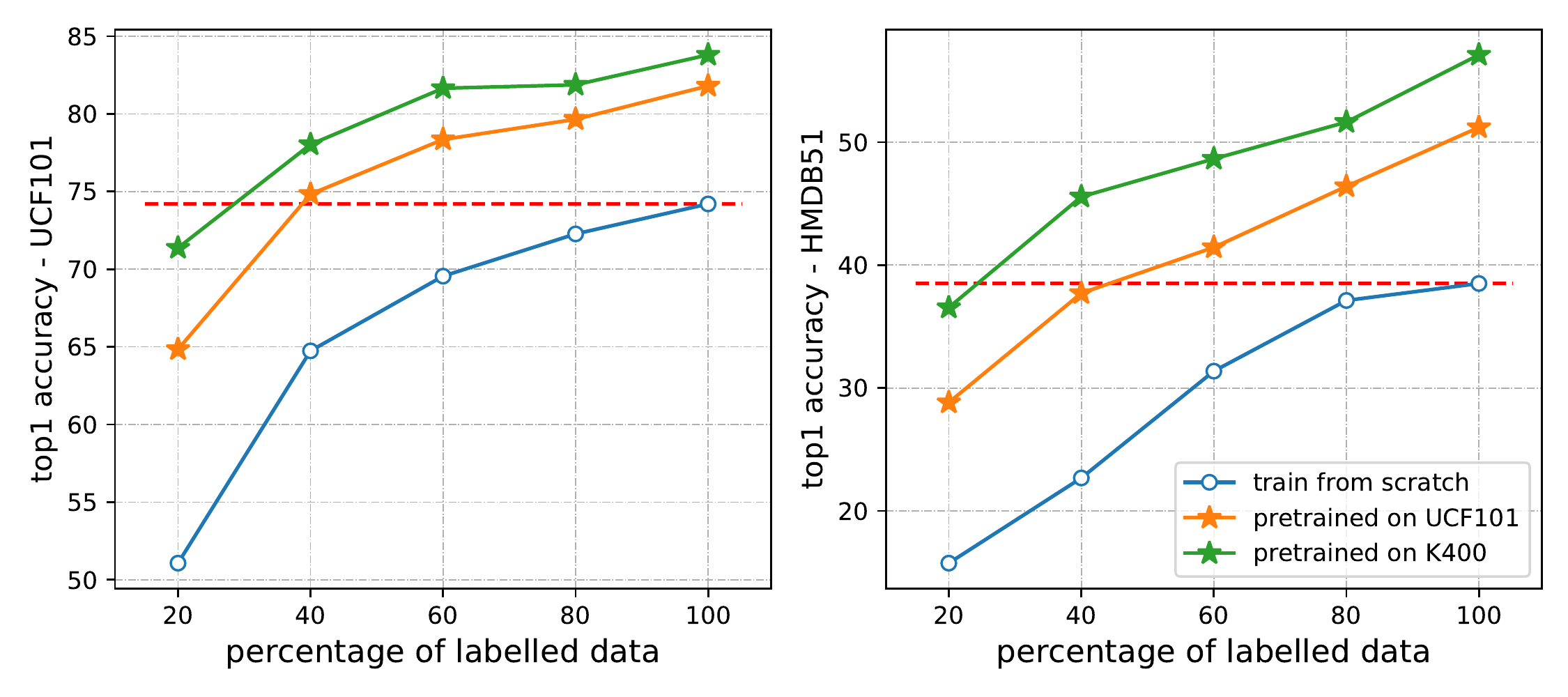}
    \caption{Data efficiency of CPNet. Models (R2+1D backbone) are pretrained with CPNet or randomly initialized, and then evaluated on action recognition task (finetuning protocol) on UCF101(L) and HMDB51(R) with different percentages of labeled data.}
    \label{fig:data_efficiency}
\end{figure}

\section{Data Efficiency}
Data efficiency are reflected by the effectiveness of the representation under a scarce-annotation regime(Han et al. 2020a). 
To show the data efficiency of the CPNet, we evaluate the pretrained model on action classification task and limit the model to only use 20\%, 40\% and 60\% and 60\% of the labelled training samples, then we report the accuracy on the same testing set.
Fig.~\ref{fig:data_efficiency} shows the action recognition testing results on UCF101 and HMDB51 datasets. 
The model's parameters are either initialized with CPNet or randomly initialized. CPNet initialization (both K400 and UCF101 pretrained) leads to considerably better results than random initialization for all percentages of training data usage. Notably, with CPNet pretraining and with only 40\% of the labeled data, we can achieve a similar (UCF101-pretrained) or a significantly better performance (K400-pretrained) than a randomly initialized classifier trained on entire training set. 

\begin{table}[t]
    \setlength{\abovecaptionskip}{0pt}
    \setlength{\belowcaptionskip}{0pt}
    \setlength{\tabcolsep}{3.6pt}
	\scriptsize
	\caption{Comparison between CRNet and CPNet on action recognition task.}

	\begin{tabular}{ccccccc}
		\hline
		Method & Pretrained & Resolution & Backbone & UCF101 & HMDB51 & Diving48 \\
		\hline
		\hline
        {CRNet} & UCF101 & 112$\times$112 & C3D & 76.7 & 46.9  & 71.5 \\
        {CRNet} & UCF101 & 112$\times$112 & R3D  & 76.9 & 47.9  & 69.8 \\
        {CRNet} & UCF101 & 112$\times$112 & R(2+1)D  & 81.7 & 51.9 & 72.0 \\
        \hline
        \textbf{CPNet} & UCF101 & 112$\times$112 & C3D & \textbf{77.5} & \textbf{45.2} & \textbf{72.6} \\
        \textbf{CPNet} & UCF101 & 112$\times$112 & R3D  & \textbf{77.2} & \textbf{46.3} & \textbf{70.1} \\
        \textbf{CPNet} & UCF101 & 112$\times$112 & R(2+1)D  & \textbf{81.8} & \textbf{51.2} & \textbf{72.8} \\
        \hline
	\end{tabular}
	\label{tab:actionclassification_supp}
\end{table}

\begin{table}[t]
    \setlength{\abovecaptionskip}{0pt}
    \setlength{\belowcaptionskip}{0pt}
    \setlength{\tabcolsep}{2.0pt}
	\centering
	\scriptsize
	\caption{Comparison between CRNet and CPNet on video retrieval task.}
	\begin{tabular}{cccccccc}
		\hline
		 \multirow{2}{*}{Backbone} &\multirow{2}{*}{Method}  & \multicolumn{5}{c}{UCF101/HMDB51} \\
		\cline{3-7} 
		 & & R@1 & R@5 & R@10 & R@20 & R@50 \\
		\hline
		\hline
      \multirow{2}{*}{C3D} & CRNet & 18.6/10.2 & 35.1/29.6 & 47.5/42.4 & 58.2/57.7 & 74.4/76.2 \\
      & \textbf{CPNet} &  \textbf{33.2/13.7} & \textbf{50.0/33.2} & \textbf{55.8/45.5} & \textbf{64.5/59.3} & \textbf{76.5/78.7} \\
      \hline
      \multirow{2}{*}{R3D} & CRNet & 23.3/12.6 & 40.6/30.7 & 51.5/42.1 & 62.7/55.4 & 76.6/75.9 \\
      & \textbf{CPNet} & \textbf{35.1/16.5} & \textbf{49.0/35.5} & \textbf{57.2/47.5} & \textbf{67.3/60.0} & \textbf{76.6/79.2} & \\
      \hline
      \multirow{2}{*}{R2+1D} &CRNet & 22.9/12.0 & 35.4/27.5 & 44.0/38.1 & 51.9/53.3 & 66.4/72.1 & \\
      & \textbf{CPNet} & \textbf{35.3/14.0} & \textbf{49.9/32.8} & \textbf{58.6/45.8} & \textbf{67.0/60.5} & \textbf{77.8/77.2} & \\
      \hline
	\end{tabular}
	\label{tab:videoretrieval_supp}
\end{table}

\section{Missing Section Estimation in RGB Space}

For the task of missing section estimation, in our method, we formulate it as a contrastive learning problem and approximate in the feature space.
We also experiment with a more straight-forward way of reconstructing the missing section in the RGB space.
As you can see in Fig. 2, we replace the third projection network, which is responsible for missing section approximation, in CPNet with a 3D-Conv decoder $G_{\theta_g}$.
The decoder takes the feature representations ($f_{{i,d}}$) of a discontinuous clip  ($c_{{i,d}}$) as input and tries to reconstruct its missing section ($c_{{i,m}}$).
We adopt the decoder structure from PRP~\cite{yao2020video} and use the mean-square-error loss as reconstruction loss $\mathscr{L}_R$ for optimization.
The other two discriminative pretext tasks of continuity justification and discontinuity localization are also reserved with two projection networks on the top of the backbone encoder. 
Like CPNet, two classification losses $\mathscr{L}_J$ and $\mathscr{L}_L$ are used for optimization.
We term this model as Continuity Reconstruction Network(CRNet).
CRNet is optimized in a joint discriminative-generative manner with the multi-loss function,

\begin{equation}
    \mathscr{L} = w_1 \times \mathscr{L}_{J}+w_2 \times \mathscr{L}_{L} + w_3\times \mathscr{L}_{R}
    \label{eq:joint-optimization2}
\end{equation}

\noindent where $w_1, w_2, w_3 \in [0,1]$ are the individual weights on the losses.
The joint optimization trains the shared backbone encoder to generate feature representations reliable for each of the pretext tasks.

To evaluate the efficacy of CRNet in spatiotemporal representation learning, we pretrain the model and evaluate on the downstream tasks of action recognition and video retrieval.
We pretrain the model on the training split of UCF101. For action recognition, we finetune and evaluate on UCF101, HMDB51, and Diving48 datasets. For video retrieval, we extract features and evaluate on UCF01 and HMDB51 datasets.
The pretraining and evaluation settings are the same as we use for CPNet.
From the results in Table~\ref{tab:actionclassification_supp}, we can observe that CRNet is inferior to CPNet on action recognition task, but the margin is small and its performance is comparable to CPNet.
From the results in Table~\ref{tab:videoretrieval_supp}, we can observe a large gap between CRNet to CPNet on video retrieval task.
CPNet wins CRNet by a large margin regardless of the backbone we use. 
About this task, we give credit to the superiority of contrastive learning mechanism over generative learning mechanism in learning more discriminative features via instance discrimination.

\end{document}